\documentclass[letterpaper, 10 pt, conference]{ieeeconf}  %

\IEEEoverridecommandlockouts
\usepackage{amsmath}

\usepackage[per-mode=fraction]{siunitx}
\usepackage[pdftex]{graphicx}
\usepackage{svg}
\graphicspath{{figures/}}
\svgpath{figures/}
\usepackage{tikz}
\usetikzlibrary{patterns, arrows, shapes, positioning}
\usepackage{pgfplots}
\pgfplotsset{compat=newest}
\usepackage{hyperref}
\hypersetup{
    colorlinks=true,
    linkcolor=black,
    citecolor=black,
    filecolor=black,
    urlcolor=black,
    breaklinks=true
}
\usepackage[T1]{fontenc}
\usepackage[utf8]{inputenc}
\usepackage{csquotes}
\usepackage[english]{babel}
\usepackage[export]{adjustbox}
\usepackage{caption}
\usepackage{subcaption}
\usepackage[colorinlistoftodos]{todonotes}
\usepackage{lipsum}
\usepackage{multirow}
\usepackage{textcomp}
\usepackage{placeins}%
\usepackage{cleveref}
\usepackage{breakurl}

\usepackage[style=ieee,
            doi=false,
            url=false,
            mincitenames=1,
            maxcitenames=1,
            minbibnames=6,
            maxbibnames=6,
            backend=biber]{biblatex}  %
\AtBeginBibliography{\footnotesize}

\title{%
Generation of Training Data from HD Maps in the Lanelet2 Framework
} 

\author{
    Fabian Immel$^{1}$,
    Richard Fehler$^{1}$,
    Frank Bieder$^{1}$ and 
    Christoph Stiller$^{2}$
    \thanks{
        $^{1}$FZI Research Center for Information Technology, Karlsruhe, Germany
        {\tt\small \{immel, fehler, bieder\}@fzi.de}
    }%
    \thanks{
        $^{2}$Institute of Measurement and Control Systems, Karlsruhe Institute of Technology (KIT),
        Karlsruhe, Germany
        {\tt\small stiller@kit.edu}
    }%
}

\begin{document}
\maketitle
\pagestyle{empty}

\begin{abstract}

    Using HD maps directly as training data for machine learning tasks has seen a massive surge in popularity and shown promising results, e.g. in the field of map perception.
    Despite that, a standardized HD map framework supporting all parts of map-based automated driving and training label generation from map data does not exist. %
    Furthermore, feeding map perception models with map data as part of the input during real-time inference is not addressed by the research community.
    In order to fill this gap, we present lanelet2\_ml\_converter, an integrated extension to the HD map framework Lanelet2, widely used in automated driving systems by academia and industry.
    With this addition Lanelet2 unifies map based automated driving, machine learning inference and training, all from a single source of map data and format. 
    Requirements for a unified framework are analyzed and the implementation of these requirements is described.
    The usability of labels in state of the art machine learning is demonstrated with application examples from the field of map perception.
    The source code is available embedded in the Lanelet2 framework under \url{https://github.com/fzi-forschungszentrum-informatik/Lanelet2/tree/feature\_ml\_converter}.

\end{abstract}
\section{Introduction}
\subsection*{HD Maps}
\label{sec:hdmaps}
For fully automated driving, an agent has to master a wide range of safety-critical tasks such as localization, prediction, behavior generation and planning.
In many of today's automated driving stacks, high-definition (HD) maps are used as an essential component in reliably solving these tasks.
By providing static contextual information, HD maps can complement a vehicle's sensor suite and online perception, which can suffer from occlusion, limited sensor range or poor weather conditions.
Hereby, HD maps typically consist out of multiple layers, which are beneficial for solving different sub-tasks of automated driving.
Due to their wide field of applications and the lack of an established map standard, HD maps can come in a wide range of formats. %
These map formats are usually tailored for one specific task or software stack and created by a specific map manufacturer, such as TomTom \cite{ONLINEtomtom} and HERE \cite{ONLINEhere}. %
Hence, map formats are hardly compatible and the corresponding software suite to create, maintain or further develop the map and its format is generally not publicly available.
\begin{figure}[htb]
    \includegraphics[width=\columnwidth]{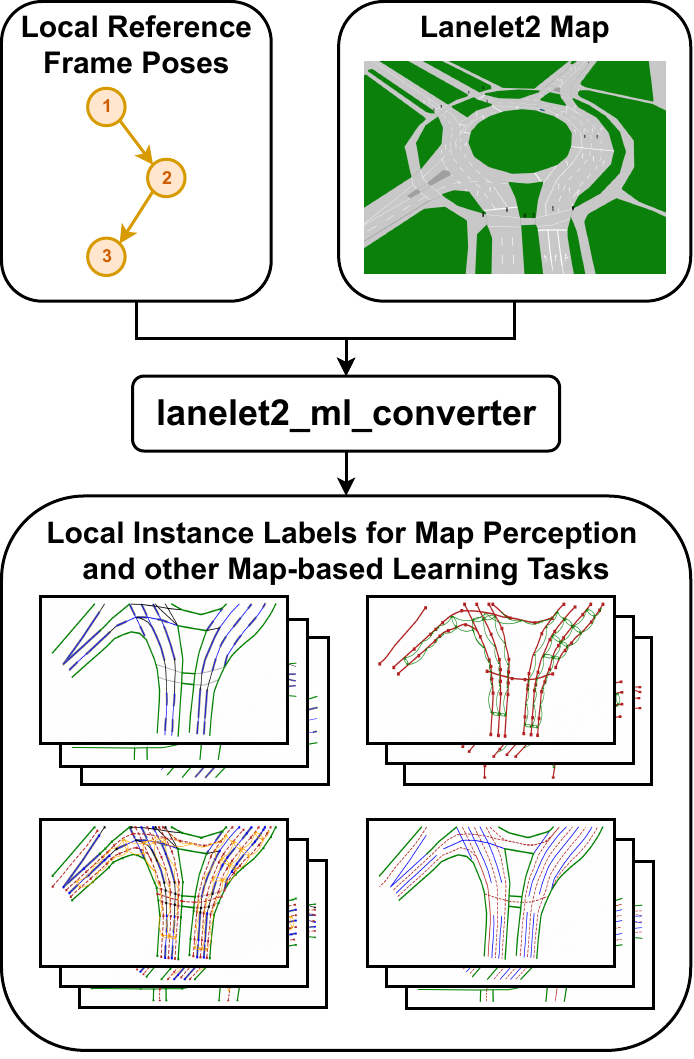}
    \caption{Usage and applications of the software module. Together with the module, a Lanelet2 map can be used as a unified format for both map-based highly automated driving and as labels for a variety of map perception and map-based learning tasks.}
    \label{fig:teaser_plot}
    \vspace{-2mm}
\end{figure}

\subsection*{Lanelet2}
\label{sec:lanelet2}
In order to fill this gap, \cite{Poggenhans2018} provides the software map framework Lanelet2.
It aims to define a unified map framework which is suitable for serving a comprehensive range of automated driving tasks.
In contrast to many other map formats, e.g. OpenDRIVE \cite{ONLINEopendrive}, Lanelet2 is modelled bottom-up which allows a modular and flexible representation of complex road topologies and intersections.
A Lanelet2 map is divided into three different layers: a physical layer containing observable objects like road markings or lane boundaries, a relational layer defining the connection between physical elements and corresponding lanes and, finally, the topological layer combining the elements of the relational layer into a drivable HD map with supporting lane-level road network. These three layers emerge bottom-up from the physical layer and form the basis of the Lanelet2 philosophy.
In addition to defining a novel map format, the Lanelet2 framework is equipped with a set of tools for simple and standardized creation of new HD maps, data retrieval, handling of map information on multiple abstraction layers and efficient real-world application in automated driving systems.
Since its release in 2018, Lanelet2 has been widely adopted and applied in various academic and industrial projects.
Support has been demonstrated for many applications such as semantic localization \cite{9341003}, HD map creation \cite{xu2022sind} and behavior planning \cite{10171371, majstorovic2023dynamic}. %
Lanelet2 is the default HD map format in most datasets related to planning and prediction tasks \cite{9304839,TheHighDDatasetaDroneDatasetofNaturalisticVehicleTrajectoriesonGermanHighwaysforValidationofHighlyAutomatedDrivingSystems,TheRounDDatasetaDroneDatasetofRoadUserTrajectoriesatRoundaboutsinGermany,TheExiDDatasetaRealWorldTrajectoryDatasetofHighlyInteractiveHighwayScenariosinGermany,xu2022sind,10171371, majstorovic2023dynamic}.
In this context the research community developed tools to convert maps into the Lanelet2 format if the supplied map is not available in this format.
As an example, the CommonRoad framework \cite{Althoff2018, commonroad2017} provides tooling for converting many map formats like openDRIVE to Lanelet2 and vice versa.
In the space of sensor datasets, \cite{Naumann2023} developed an automatic approach to enhance the existing HD road maps of nuScenes and convert them into the Lanelet2 format offering a comprehensive HD map database including the full road topology.
The extensive functionalities of Lanelet2, efficient implementation and ROS 1/2 support \cite{ROS2} enable most modern academic and open-source automated driving stacks \cite*{autoware-doc}. %

\subsection*{Online Map Perception}
\label{sec:new-map-learning}

Map perception is one of the key technologies in scaling automated driving.
Online map perception can compensate outdated or incomplete map data, extending the available geolocations of automated driving systems and offline map perception can support cloud mapping and map updates.
Training of map perception models is feasible for academic research, thanks to the release of several large-scale sensor datasets \cite{av2_trust_but_verify,ScalabilityinPerceptionforAutonomousDrivingWaymoOpenDataset,wilson2023argoverse,nuscenes,openlaneV2}, that include either city-wide or scenario-specific maps on a lane-level with corresponding sensor data.
This accelerates the development of methods for perception of low level map features such as road boarders, lane dividers etc., as well as higher level relationships of map components such as the assignment of traffic elements to lanes.
Lanelet2 maps have been an established format for motion prediction datasets, and now increasingly become available for large scale sensor datasets \cite*{Naumann2023}.
Usual maps released with sensor datasets are published in a custom format, with varying fidelity and map content without a standardized API and format.
These maps and supporting software have not demonstrated the ability to run classic map based automated driving software stacks such as Autoware \cite{autoware-doc} and do not implement or support features such as path planning, map validation or routing. %

\subsection*{Contributions}

In the following we summarize the main contributions of this work:
\begin{itemize}
    \item We motivate the need for a unified HD map format and framework that is flexible enough to be used in fully map based driving, can derive map information of different abstraction levels and thus allows for training map perception models.
    \item We define a set of requirements for a framework that is suitable for generating training data from HD maps. %
    \item Embedded in the Lanelet2 library, we design an extension of the framework that follows this set of requirements. We describe implementation details and provide source-code. %
    \item We demonstrate how the lanelet2\_ml\_converter extension generates training data from Lanelet2 maps for a variety of recent map perception problems in different abstraction layers. %
\end{itemize}

\section{Requirements for Modern HD Map Frameworks}
\label{sec:req}

In this section we specify the requirements needed to exceed the state of the art provided by \cites{Poggenhans2018}{ONLINEopendrive} and especially datasets like nuScenes \cite{nuscenes} or Argoverse 2 \cite{Argoverse2}, for online map perception and learning from 3D map labels.

\subsection*{Unified HD Map Framework for Automated Driving and Deep Learning Inference and Training}
To implement state of the art map perception in automated driving systems, a HD map framework needs to support training data generation as well as fast online local instance label retrieval in the necessary inference representation.
Additionally support for classical tasks like manual and automatic mapping, map updates, behavior generation and path planning are necessary to build a automated driving system and can not be disregarded.
\cite{maptrv2}\cite{openlaneV2} use map formats, that lack the expressive power necessary to enable automated driving in complex operational domains and do not provide software support for the aforementioned classical tasks.
Local instance label retrieval methods are implemented by \cite{maptrv2}\cite{openlaneV2}, but their high runtime latency and a generally offline oriented software framework hinder progress in real-world application of these methods.

\subsection*{Generation of Training Labels from Maps}
These local instance labels are used in offline applications like scalable label generation for perception, given a high-quality localization.
Ideally these local instances include not only physically present features such as road markings, but also high level information such as yielding rules, the lane-graph structure given by the relationship to lane successors and neighbors, driving directions and relationships to traffic elements as signs and lights.
Depending on the learning task, different parts of the map and the corresponding learning representation can be loaded.

\subsection*{Traceability}
In case of corrupted or false instance labels, it is necessary to backtrace them to their origin in the HD map.
This demands a traceable connection between a instance label and the corresponding map element it is derived from.
This can be extended up to the underlying mapping process of the map element, allowing for full quality control.

\subsection*{Map Validation} %
HD Maps included in publicly available automated driving datasets lack the provided software API or formal definition to check for validity regarding traffic rule compliance and traffic flow consistency.
A map validation module within the map framework can ensure a more meaningful, consistent and less error-prone label generation.
This is especially crucial for the generation of higher-level training labels, which are inferred from multiple interacting map annotations and are sensitive to incomplete or inconsistent map annotations.
A well tested system for this is provided by the Lanelet2 validation software module \cite{Poggenhans2018}, however it is not usable for map labels without a process to generate labels directly from Lanelet2 maps, like in the extension proposed in this paper.
This validation additionally supports automatized mapping frameworks as in the work of \cite{ll2_mapping}.

\subsection*{Independence of Label Instances from Map Annotation Artifacts}
Both manual and automatic mapping can produce almost infinitely many permutations of map element annotations, with different polyline point positions, ordering, breaks in polylines, and resulting lane subdivisions into lanelets.
All these annotation permutations can represent the very same objects and information in the real world.
To ensure consistency, a unambiguous representation of the corresponding instance label needs to be derivable for map elements.
This requires a deterministic solving of map annotation artifacts or permutations provoked by the annotation process.

\subsection*{Support for Varying Local Reference Frame Poses}

Different datasets and mapping setups, produce reference frame poses with varying degrees of freedom. Training on custom setups and public datasets requires support for varying pose qualities.

\subsection*{Real-Time Capability} %
While the generation of training data is usually performed offline, an online application of the new module would be required to fuse the map perception with prior knowledge given by the HD map.
This would require a high-performant implementation regarding the processing of selected map elements and their conversion to efficient numerical representations.
\section{Implementation}
\label{sec:implementation}
This chapter describes the extension of the Lanelet2 framework implementing the requirements specified in \cref{sec:req}.
\begin{figure}[!htb]
    \centering
    \begin{subfigure}[c]{0.48\columnwidth}
        \includegraphics[width=\columnwidth]{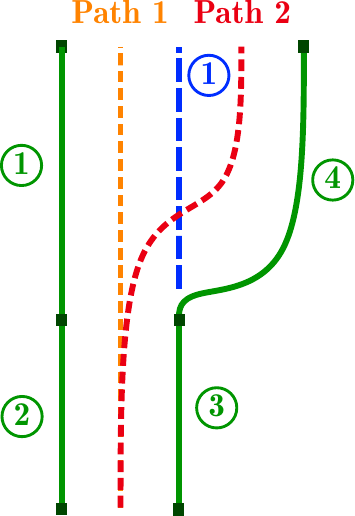}
        \caption{Original map labels}
        \label{fig:compound_label_explanation_1}
        \vspace{3mm}
    \end{subfigure}
    \begin{subfigure}[c]{0.48\columnwidth}
        \includegraphics[width=\columnwidth]{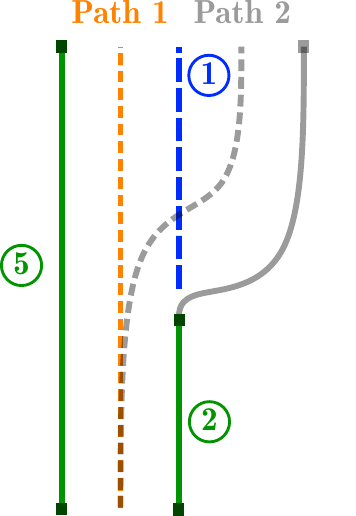}
        \caption{Compound feat. of path 1}
        \label{fig:compound_label_explanation_2}
        \vspace{3mm}
    \end{subfigure}
    \begin{subfigure}[c]{0.48\columnwidth}
        \centering
        \includegraphics[width=\columnwidth]{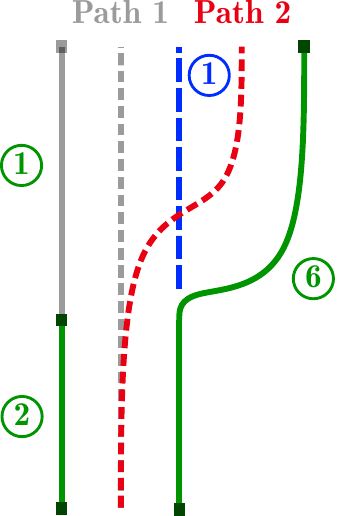}
        \caption{Compound feat. of path 2}
        \label{fig:compound_label_explanation_1}
    \end{subfigure}
    \begin{subfigure}[c]{0.48\columnwidth}
        \centering
        \includegraphics[width=\columnwidth]{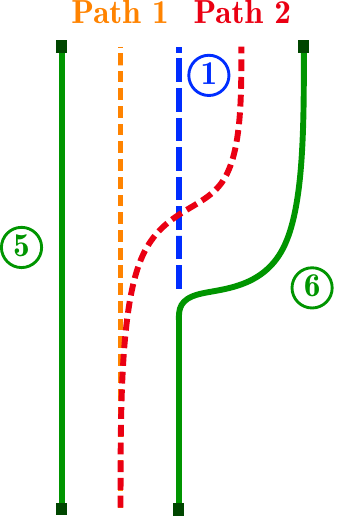}
        \caption{Overlapping labels deleted}
        \label{fig:compound_label_explanation_2}
    \end{subfigure}
    \caption{Visualization of compound labels and the elimination of overlaps. The overlapping compound road borders 2 and 4 are eliminated in favor of the larger compound road borders 5 and 6.}
    \label{fig:compound_label_explanation}
    \vspace{-2mm}
\end{figure}

\subsection*{Compound Labels for Independence from Map Annotation Artifacts}

To implement the essential requirement of independence from map annotation artifacts, both local and non-local,
we make use of a concept we name "compound labels".
The underlying idea is used in many online HD map construction models (e.g. \cite{maptr, vectormapnet}) and replaces each coherent chain of individual polylines with one single polyline. %
Lanelet2 and most other map formats divide the road into smaller subsegments, often referred to as lanelets.
While this has many advantages like map consistency and non-duplication, it also causes issues with local instance label generation for map learning.
The subsegmentation of labels introduces an unnecessary source of complexity in the labels and the cropping at the borders of the region of interest changes the representation of a map element in each local instance label.

Many individual polylines thus introduced form a coherent chain of the same polyline type and can be replaced by a single line, greatly simplifying the learning task for the HD map construction model.
Following the commonly used types in the area, compound labels are separated between road borders, lane dividers and lane centerlines.
\subsection*{Compound Label Generation}

The consistent and universally applicable generation of these compound labels encompasses a significant part of the implementation and is described here in greater detail.
In general, the algorithm uses the routing graph from \cite{Poggenhans2020} of the region of interest submap to extract all possible lanelet paths in the submap and follows those paths to merge the individual lanelet boundaries into compound labels.

The path following is executed by a recursive algorithm, which is initialized by the set of lanelets without a predecessor in the routing graph of the local submap.
A path of such a lanelets successors is followed until a lanelet no longer has any successors, recursively branching for a lanelet which has multiple successors.
These branching paths follow the same pattern, but are initialized with the already before followed path.
Figure \ref{fig:compound_label_explanation} shows the paths generated in this fashion for an example road segment.

Each path is then followed again and a compound label is composed of the left and right boundaries of the path lanelets as well as their centerlines, splitting at changes of the boundary type for the boundary labels.
As illustrated by Figure \ref{fig:compound_label_explanation}, the right border of path 1 results in two compound labels due to the change from road border to dashed line.

Since the same lanelet can be passed through by multiple paths, multiple compound labels containing the same map element can be created. %
This is also the case for lane borders shared by two adjacent lanelets.
A map polyline can only ever be part of one compound label, so we can eliminate all but one of the overlapping compound labels.
This is also shown in Figure \ref{fig:compound_label_explanation}, where the smaller compound road borders 2 and 4 of path 1 and 2 are eliminated in favor of the larger compound road borders 5 and 6, also from path 1 and 2 respectively.
The elimination of overlapping labels is not performed for the lane centerlines, as the underlying assumption is not applicable in this case and the training with independent path labels has been shown to improve performance \cite{lanegap}. %

All local instance labels are cropped with respect to the region of interest and resampled with a fixed number of points. If this is not needed by the application, partially processed instance labels can be generated, for example with map polyline sampling.

\subsection*{Traceability}

\begin{figure}[!htb]
    \begin{subfigure}[c]{\columnwidth}
        \centering
        \includegraphics[width=\columnwidth]{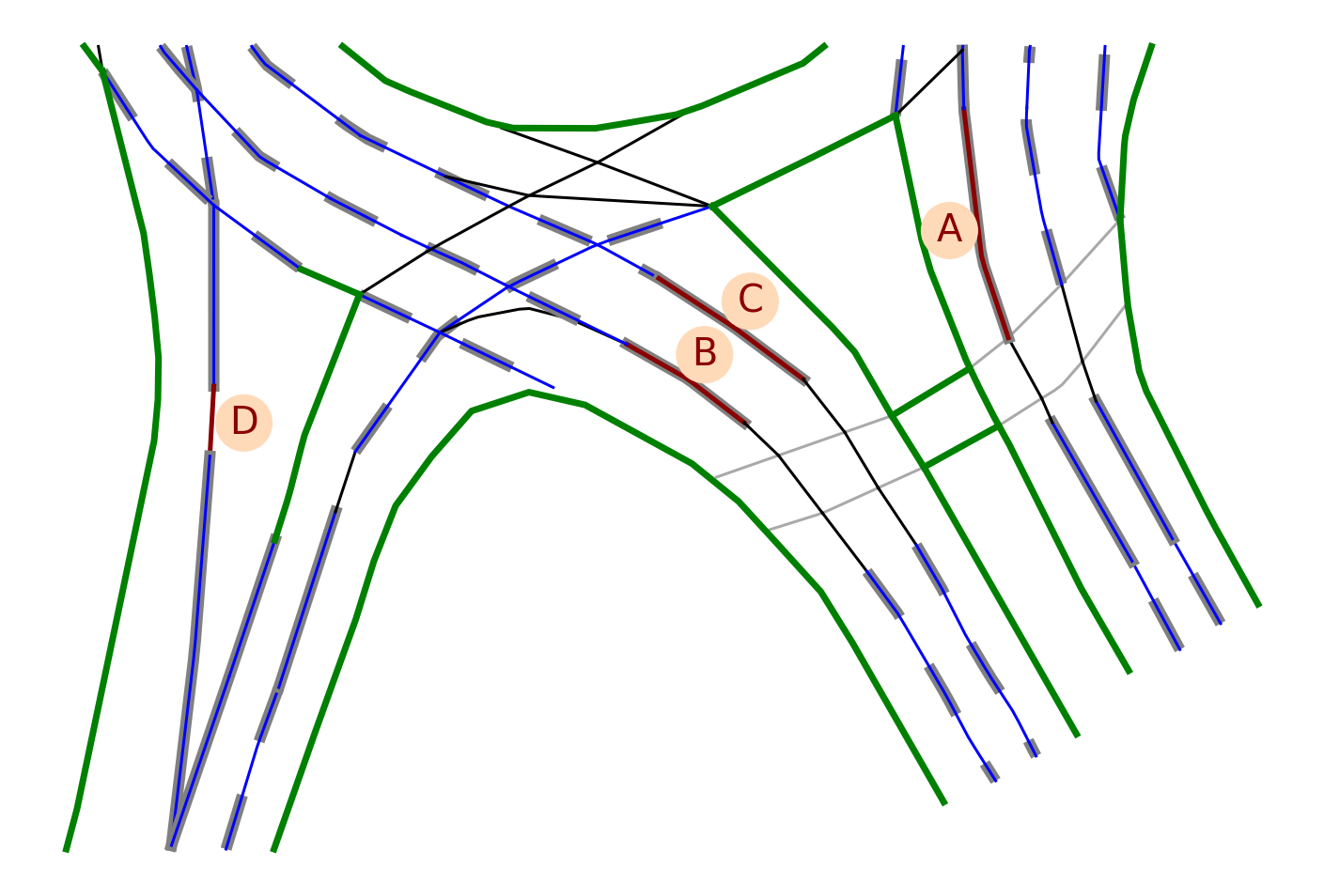}
        \caption{Original local instance labels}
        \label{fig:traceability_original}
    \end{subfigure}
    \begin{subfigure}[c]{\columnwidth}
        \centering
        \includegraphics[width=\columnwidth]{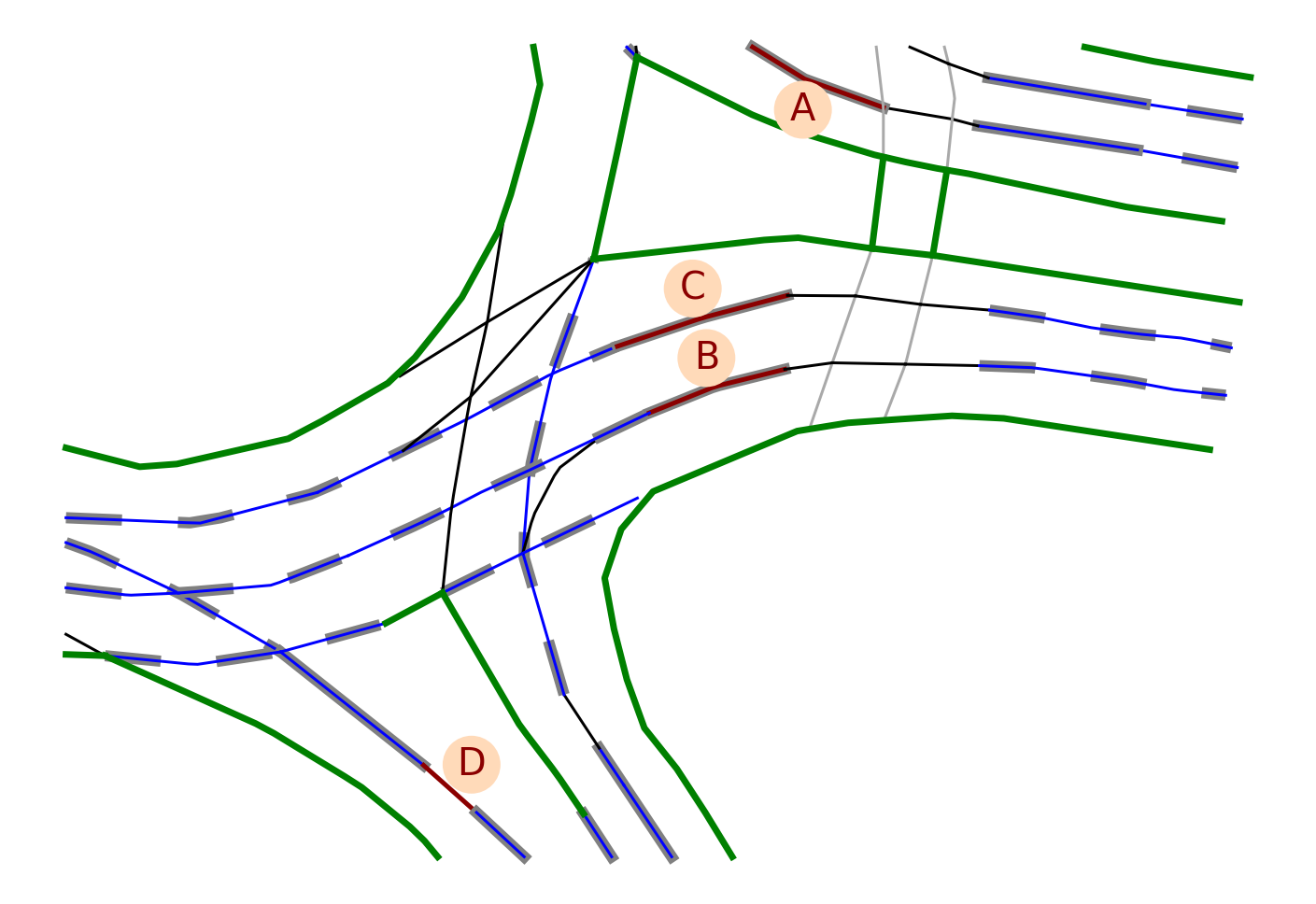}
        \caption{Different local instance labels}
        \label{fig:traceability_different}
    \end{subfigure}
    \caption{Traceability of labels in local instance labels. Every label is associated with the underlying map element, visualized here with the letters A to D. This association is maintained over local instance labels. Compound labels have traceability of the composing labels available as well.}
    \label{fig:traceability_visualization}
    \vspace{-2mm}
\end{figure}

A key requirement for ground truth generation in scale is the traceability of labels to the origin of their data, which in our case is the underlying Lanelet2 map.
This requirement is fulfilled here using the unique ID of each map element in a Lanelet2 map.
The ID of the underlying map element is available for every local instance label and carried through all processing steps, preserving traceability across geometric operations.
An example can be seen in \ref{fig:traceability_visualization}, where labels A-D are uniquely identified across different local reference frames based on their associated map element ID.

The association with map elements is preserved for compound labels as well, which contain the IDs of the map elements they are composed of as well as the arc lengths of the individual labels in the compounded polyline.
Centerline labels are associated to the respective lanelet, as no explicit geometric primitive counterpart exists in a lanelet map.

\subsection*{Support for Varying Local Reference Frame Poses}

lanelet2\_ml\_converter supports 3D reference frame poses composed of a 2D position on a flat plane and heading of the vehicle as the minimum required input.
To generate the highest quality of local instance labels, up to full 6D reference frame poses can be used.

\subsection*{Real-Time Capable Implementation}

To enable online inference of models that use map data as an input, not just as training labels, a real time capable implementation is necessary.
The extension is therefore realized in a highly performant C++ implementation, averaging a single threaded performance of 3 ms for generating a set of local instance labels from a map, including conversions to learning representations, on a Intel i7-11800H processor. %
\section{Application examples}
\label{sec:application}

\begin{figure*}[htb]
    .\centering
    \includegraphics[width=0.8\linewidth]{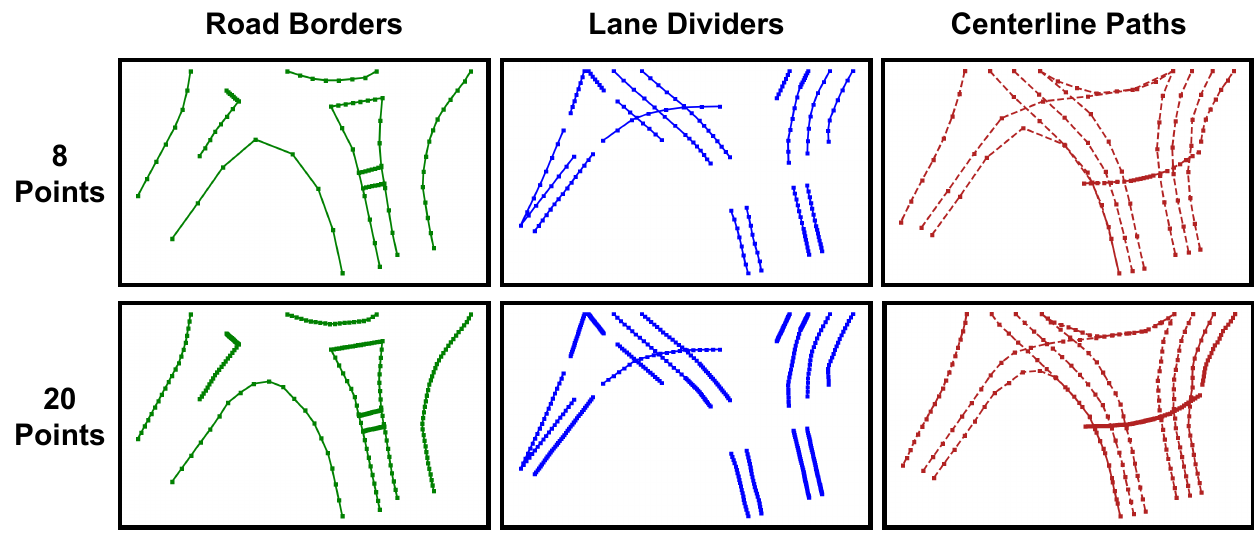}
    \caption{Labels for road borders, lane dividers and lane centerline paths with different fixed point numbers per feature similar to the labels used in MapTR \cite{maptr} and MapTRV2 \cite{maptrv2}.}
    \label{fig:maptr_labels_plot}
    \vspace{-2mm}
\end{figure*}
The map instance labels supplied from the extension can be used to supervise a wide variety of map perception training tasks. As described in \cref{sec:implementation}, its low latency allows using map data as input during inference of deployed models. 
Besides perception, other map-based learning tasks using map data as input such as trajectory prediction can be fed by this work.
In the following we demonstrate the flexibility of the resulting local label instances with three popular application examples from the field of map perception.

\subsection*{MapTR: Online HD Map Construction}
One of the key features of the recent MapTR architecture for vectorized map perception is the representation of map elements as compounded linestrings with a fixed point number.
This is natively supported in the extension and all element types can be separately used and easily resampled with a chosen fixed point number.
Figure \ref{fig:maptr_labels_plot} shows the default element types with both a 8-point and 20-point representation.

\subsection*{OpenLaneV2: Topology Inference}

\begin{figure}[htb]
    \includegraphics[width=\columnwidth]{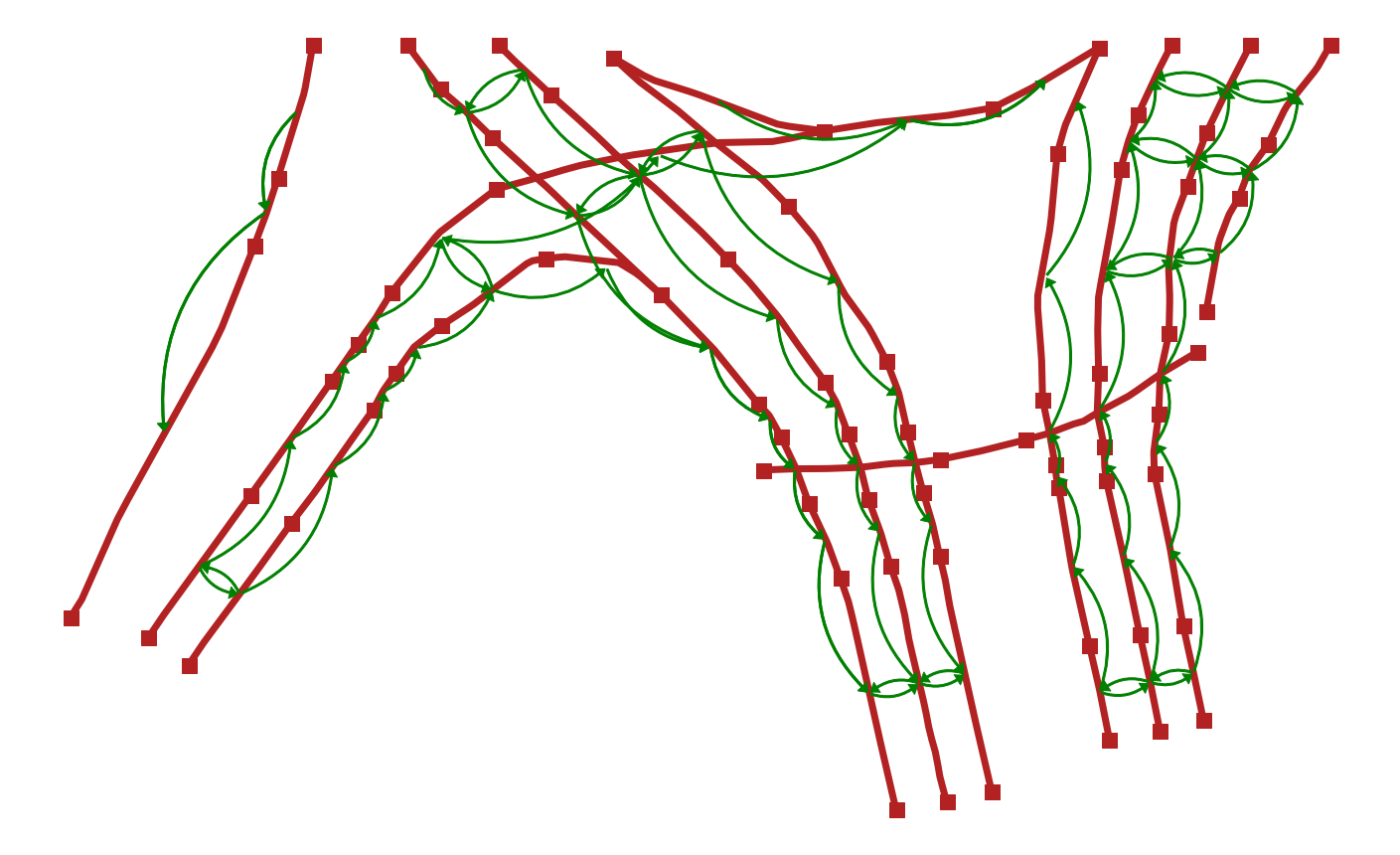}
    \caption{Labels for lanelet centerlines and their connectivity to successors and neighbours visualized as green edges similar to the OpenLaneV2 dataset \cite{openlaneV2}.}
    \label{fig:openlane_labels_plot}
    \vspace{-2mm}
\end{figure}

Another important area in map perception is scene topology inference, the inference of higher level information such as lane connectivity and traffic element to lane association.
Topology inference has seen a large boost in research activity from the OpenLaneV2 dataset \cite{openlaneV2}, a new large-scale dataset that explicitly includes higher level labels of this kind.
The label representation used in OpenLaneV2 can easily be mirrored from the label instances in the extension, with both lanelet centerlines
and their connectivity edges already part of the label data.
A visualization of the formed graph can be seen in figure \ref{fig:openlane_labels_plot}.

\subsection*{Online Map Inference and Fusion}

The full traceability of all labels to their underlying map elements and the real time capability of the extension enable another application so far unsupported by existing map data converters: Online map fusion of incomplete maps or updating existing map elements\cite*{MindtheMapAccountingforExistingMapInformationWhenEstimatingOnlineHDMapsfromSensorData}.
The Argoverse 2 dataset defines a similar task with the trust-but-verify subset \cite{av2_trust_but_verify} and this extension bridges the gap towards applicability in a real system using a unified, standardized HD map framework in all parts of an automated driving stack.
\section{Conclusions and Future Work}
\label{sec:con}

This work describes a extension to the Lanelet2 framework that enables data generation for online map perception in automated driving.
The current state of HD map formats and frameworks is reviewed and based on new developments in HD maps and map perception, the need for a unified HD map format and framework that can enable both fully map based driving and map perception tasks is motivated.
Our proposed requirements for such a framework are described together with limitations of currently available alternatives.
We detail the implementation of these requirements by this Lanelet2 extension and finally show application examples for recent popular map perception tasks.
Future work could include improved support for non-road features such as traffic lights and traffic signs.
\section*{Acknowledgements}

We would like to thank our research partner Mercedes-Benz AG for the fruitful collaboration.

\printbibliography

\end{document}